# INTERNAL JOINT FORCES IN DYNAMICS OF A 3-*PRP* PLANAR PARALLEL ROBOT

Stefan STAICU [1], Damien CHABLAT [2]

[1] Department of Mechanics, University "Politehnica" of Bucharest, Romania
[2] Institut de Recherche en Communications et Cybernétique de Nantes, France
staicunstefan@yahoo.com        Damien.Chablat@irccyn.ec-nantes.fr

Recursive matrix relations for the complete dynamics of a 3-*PRP* planar parallel robot are established in this paper. Three identical planar legs connecting to the moving platform are located in the same vertical plane. Knowing the motion of the platform, we develop first the inverse kinematical problem and determine the positions, velocities and accelerations of the robot. Further, the inverse dynamic problem is solved using an approach based on the principle of virtual work. Finally, some graphs of simulation for the input powers of three actuators and the internal joint forces are obtained.

*Key words*:  Kinematics, Dynamics, Planar parallel robot, Virtual work

## 1. INTRODUCTION

Parallel manipulators are closed-loop mechanisms that consist of separate serial chains connecting the fixed base to the moving platform. Compared with serial manipulators, the followings are the potential advantages of parallel architectures: higher kinematical precision, lighter weight and better stiffness, greater load bearing, stabile capacity and suitable position of arrangement of actuators. Equipped with revolute or prismatic actuators, parallel manipulators have a robust construction and can move bodies of large dimensions with high velocities and accelerations. That is reason why the devices, which produce translation or spherical motion to a platform, technologically are based on the concept of parallel manipulators [1].

Over the past two decades, parallel manipulators have received more attention from researches and industries. Important companies such as Giddings & Lewis, Ingersoll, Hexel and others have developed them as high precision machine tools. Considerable efforts have been devoted to the kinematics and dynamic analysis of fully parallel manipulators. Among these, the class of manipulators known as Stewart-Gough platform focused great attention (Stewart [2]; Merlet [3]). They are used in flight simulators and more recently for Parallel Kinematics Machines. The prototype of Delta parallel robot (Clavel [4]; Tsai and Stamper [5]) developed by Clavel at the Federal Polytechnic Institute of Lausanne and by Tsai and Stamper at the University of Maryland as well as the Star parallel manipulator (Hervé and Sparacino [6]) are equipped with three motors, which train on the mobile platform in a three-degrees-of-freedom general translation motion. Angeles [7], Wang and Gosselin [8] analysed the kinematics, dynamics and singularity loci of Agile Wrist spherical robot with three actuators.

Planar parallel robots are useful for manipulating an object on a plane. A mechanism is said to be a *planar robot* if all the moving links in the mechanism perform the planar motions. For a planar mechanism, the loci of all points in all links can be drawn conveniently on a plane. In a planar linkage, the axes of all revolute joints must be normal to the plane of motion, while the direction of translation of a prismatic joint must be parallel to the plane of motion. Bonev, Zlatanov and Gosselin [9] describe several types of singular configurations by studying the direct kinematics model of a 3-*RPR* planar parallel robot with actuated base joints. Aradyfio and Qiao [10] examined the inverse kinematics solution for the three different 3-DOF planar parallel robots, while Pennock and Kassner [11] present a kinematical study of a planar parallel robot where a moving platform is connected to a fixed base by three links, each leg consisting of two binary links and three parallel revolute joints. Sefrioui and Gosselin [12] give an interesting numerical solution in the inverse and direct kinematics of this kind of planar robot.

Using dual-number quaternion algebra, Mohammadi-Daniali et al. [13], [14] present a study of velocity relationships and singular conditions for general planar parallel robots.

## 2. KINEMATICS ANALYSIS

A recursive method is introduced in the present paper, to reduce significantly the number of equations and computation operations by using a set of matrices for kinematics and complete dynamics of the 3-*PRP* planar parallel robot. Having a closed-loop structure, this robot is a symmetrical mechanism composed of three planar kinematical chains with identical topology, all connecting the fixed base to the platform. The points $A_0$, $B_0$, $C_0$ represent the summits of a triangular base and other three points define the geometry of the moving platform. Each leg consists of two links, with one revolute and two prismatic joints. The parallel mechanism with seven links $(T_k, k=1,2,...,7)$ consists of three revolute joint and six prismatic joints (Fig. 1).

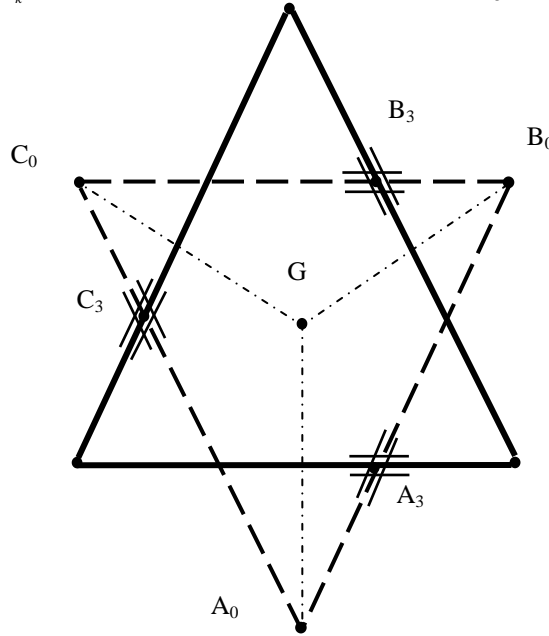

Fig. 1 The 3-*PRP* planar parallel robot

In the actuation scheme *PRP* each prismatic joint is an actively controlled prismatic cylinder. Thus, all prismatic actuators can be installed on the fixed base. For the purpose of analysis, we attach a Cartesian frame $Ox_0y_0z_0(T_0)$ to the fixed base with its origin located at triangle centre $O$, the $z_0$ axis perpendicular to the base and the $x_0$ axis pointing along the direction $C_0B_0$. Another mobile reference frame $Gx_Gy_Gz_G$ is attached to the moving platform. The origin of this coordinate central system is located just at the centre $G$ of the moving triangle. In what follows we consider that the moving platform is initially located at a *central configuration*, where the platform is not rotated with respect to the fixed base and the mass centre $G$ is at the origin $O$ of the fixed frame. It is noted that the relative rotation of $T_k$ body with $\varphi_{k,k-1}$ angle must be always pointing about the direction of $z_k$ axis.

One of three active legs (for example leg $A$) consists of a prismatic joint, which is as well as a piston **1** of mass $m_1$ linked at the $A_1x_1^Ay_1^Az_1^A$ frame, having a rectilinear motion of displacement $\lambda_{10}^A$. Second element of the leg is a rigid body **2** linked at the $A_2x_2^Ay_2^Az_2^A$ frame, having a relative rotation about $z_2^A$ axis with the angle $\varphi_{21}^A$. It has the mass $m_2$ and tensor of inertia $\hat{J}_2$ with respect to $T_2^A$ frame. Finally, a prismatic joint is introduced at a planar moving platform as an equilateral triangle with the edge $l = l_0\sqrt{3}$, mass $m_3$ and inertia tensor $\hat{J}_3$ with respect to $A_3$, which translate relatively along $z_3^A$ axis with the displacement $\lambda_{32}^A$ (Fig. 2).

At the central configuration, we also consider that all legs are symmetrically extended and that the angles of orientation of three edges of fixed platform are given by

$$\alpha_A = \frac{\pi}{3}, \alpha_B = \pi, \alpha_C = -\frac{\pi}{3}. \tag{1}$$

In the study of the kinematics of robot manipulators, we are interested in deriving a matrix equation relating the location of an arbitrary body to the joint variables. We call the matrix $a^{\varphi}_{k,k-1}$, for example, the orthogonal transformation $3\times 3$ matrix of relative rotation with the angle $\varphi^A_{k,k-1}$ of link $T^A_k$ around $z^A_k$ axis. Starting from the origin $O$ and pursuing three independent legs $OA_0A_1A_2A_3$, $OB_0B_1B_2B_3$, $OC_0C_1C_2C_3$, we obtain the following transformation matrices [15]

$$q_{10} = \theta_1\theta^i_\alpha,\ q_{21} = q^\varphi_{21}\theta^T_1,\ q_{32} = \theta_1\theta_2 \quad (q = a, b, c) \quad (i = A, B, C), \tag{2}$$

where

$$\theta^i_\alpha = \begin{bmatrix} \cos\alpha_i & \sin\alpha_i & 0 \\ -\sin\alpha_i & \cos\alpha_i & 0 \\ 0 & 0 & 1 \end{bmatrix}, \theta_1 = \begin{bmatrix} 0 & 0 & -1 \\ 0 & 1 & 0 \\ 1 & 0 & 0 \end{bmatrix}, \theta_2 = \frac{1}{2}\begin{bmatrix} 1 & -\sqrt{3} & 0 \\ \sqrt{3} & 1 & 0 \\ 0 & 0 & 2 \end{bmatrix}$$

$$q^\varphi_{k,k-1} = rot(\varphi^i_{k,k-1}) = \begin{bmatrix} \cos\varphi^i_{k,k-1} & \sin\varphi^i_{k,k-1} & 0 \\ -\sin\varphi^i_{k,k-1} & \cos\varphi^i_{k,k-1} & 0 \\ 0 & 0 & 1 \end{bmatrix},\ q_{k0} = \prod_{s=1}^{k} q_{k-s+1,k-s} \quad (k=1,2,3). \tag{3}$$

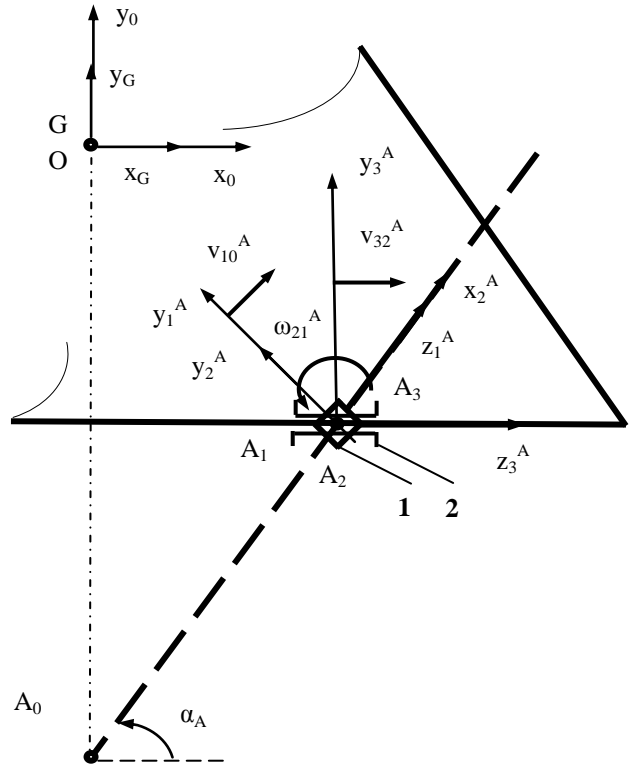

Fig. 2 Kinematical scheme of first leg $A$ of the mechanism

In the inverse geometric problem, it can be considered that the position of the mechanism is completely given through the coordinates $x^G_0$, $y^G_0$ of the mass centre $G$ of the moving platform and the orientation angle $\phi$ of the mobile central frame $Gx_Gy_Gz_G$. The orthogonal known rotation matrix of the platform from $Ox_0y_0z_0$ to $Gx_Gy_Gz_G$ reference system is $R = rot(\phi)$. We suppose that the position vector of $G$ centre $\vec{r}^G_0 = [x^G_0\ y^G_0\ 0]^T$ and the orientation angle $\phi$, which are expressed by following analytical functions

$$\frac{x_0^G}{x_0^{G*}} = \frac{y_0^G}{y_0^{G*}} = \frac{\phi}{\phi^*} = 1 - \cos\frac{\pi}{3}t \tag{4}$$

can describe the general absolute motion of the moving platform in its *vertical plane*. From the conditions concerning the orientation of the platform

$$q_{30}^{\circ T} q_{30} = R, \quad (q = a, b, c) \tag{5}$$

with

$$q_{30} = q_{32} q_{21} q_{10}, \quad q_{30}^{\circ} = \theta_1 \theta_2 \theta_\alpha^i, \quad (i = A, B, C), \tag{6}$$

we obtain the first following relations between angles

$$\varphi_{21}^A = \varphi_{21}^B = \varphi_{21}^C = \phi. \tag{7}$$

Six independent variables $\lambda_{10}^A, \lambda_{32}^A, \lambda_{10}^B, \lambda_{32}^B, \lambda_{10}^C, \lambda_{32}^C$ will be determined by several vector-loop equations as follows

$$\vec{r}_{10}^i + \sum_{k=1}^{2} q_{k0}^T \vec{r}_{k+1,k}^i + q_{30}^T \vec{r}_3^{Gi} = \vec{r}_0^G \quad (q = a, b, c) \ (i = A, B, C), \tag{8}$$

where

$$\vec{r}_{10}^i = \vec{r}_{00}^i + (l_0/\sqrt{3} + \lambda_{10}^i) q_{10}^T \vec{u}_3, \ \vec{r}_{21}^i = \vec{0}, \ \vec{r}_{32}^i = \lambda_{32}^i q_{32}^T \vec{u}_3, \ \vec{r}_3^{Gi} = 0.5l_0 [0 \quad 1 \quad -1/\sqrt{3}]^T$$

$$\vec{r}_{00}^A = l_0 [0 \quad -1 \quad 0]^T \ \vec{r}_{00}^B = 0.5l_0 [\sqrt{3} \quad 1 \quad 0]^T, \ \vec{r}_{00}^C = 0.5l_0 [-\sqrt{3} \quad 1 \quad 0]^T$$

$$\vec{u}_1 = \begin{bmatrix} 1 \\ 0 \\ 0 \end{bmatrix}, \vec{u}_2 = \begin{bmatrix} 0 \\ 1 \\ 0 \end{bmatrix}, \vec{u}_3 = \begin{bmatrix} 0 \\ 0 \\ 1 \end{bmatrix}, \tilde{u}_3 = \begin{bmatrix} 0 & -1 & 0 \\ 1 & 0 & 0 \\ 0 & 0 & 0 \end{bmatrix}. \tag{9}$$

Actually, these vector equations mean that

$$(\frac{l_0}{\sqrt{3}} + \lambda_{10}^i)\cos\alpha_i + \lambda_{32}^i \cos(\phi - \frac{\pi}{3} + \alpha_i) = x_0^G - x_{00}^i - \frac{l_0}{2\sqrt{3}}\cos(\phi + \alpha_i) + \frac{l_0}{2}\sin(\phi + \alpha_i)$$

$$(\frac{l_0}{\sqrt{3}} + \lambda_{10}^i)\sin\alpha_i + \lambda_{32}^i \sin(\phi - \frac{\pi}{3} + \alpha_i) = y_0^G - y_{00}^i - \frac{l_0}{2\sqrt{3}}\sin(\phi + \alpha_i) - \frac{l_0}{2}\cos(\phi + \alpha_i) \quad (i = A, B, C). \tag{10}$$

The rotations of the compounding elements of each leg (for example the leg *A*) are characterized by recursive relations of following skew-symmetric matrices

$$\tilde{\omega}_{k0}^A = a_{k,k-1} \tilde{\omega}_{k-1,0}^A a_{k,k-1}^T + \omega_{k,k-1}^A \tilde{u}_3, \ \omega_{k,k-1}^A = \dot{\varphi}_{k,k-1}^A, \ (k = 1,2,3) \tag{11}$$

which are *associated* to the absolute angular velocities

$$\vec{\omega}_{10}^A = \vec{0}, \vec{\omega}_{21}^A = \dot{\phi}\vec{u}_3, \vec{\omega}_{32}^A = \vec{0}, \vec{\omega}_{20}^A = \dot{\phi}\vec{u}_3, \vec{\omega}_{30}^A = \dot{\phi}\vec{u}_3. \tag{12}$$

Following relations give the velocities $\vec{v}_{k0}^A$ of the joints $A_k$

$$\vec{v}_{10}^A = \dot{\lambda}_{10}^A \vec{u}_3, \vec{v}_{21}^A = \vec{0}, \vec{v}_{32}^A = \dot{\lambda}_{32}^A \vec{u}_3, \vec{v}_{k0}^A = a_{k,k-1} \vec{v}_{k-1,0}^A + a_{k,k-1} \tilde{\omega}_{k-1,0}^A \vec{r}_{k,k-1}^A + v_{k,k-1}^A \vec{u}_3. \tag{13}$$

Equations of geometrical constraints (7) and (8) when differentiated with respect to time lead to the following *matrix conditions of connectivity* [16]

$$v_{10}^A \vec{u}_j^T a_{10}^T \vec{u}_3 + v_{32}^A \vec{u}_j^T a_{30}^T \vec{u}_3 = \vec{u}_j^T \dot{\vec{r}}_0^G - \dot{\phi} \vec{u}_j^T a_{20}^T \tilde{u}_3 \{\vec{r}_{32}^A + a_{32}^T \vec{r}_3^{GA}\} \quad (j = 1, 2). \tag{14}$$

From these equations, we obtain the relative velocities $v_{10}^A, \omega_{21}^A, v_{32}^A$ as functions of angular velocity of the platform and velocity of mass centre *G*. The Jacobian matrix given by these conditions of constraints is a fundamental element for the analysis of the robot workspace and the configurations of singularities where the manipulator becomes uncontrollable [17].

Concerning the first leg *A*, the characteristic *virtual velocities* are expressed as functions of the pose of the mechanism by the general kinematical equations (14), where we add the contributions of successive virtual translations during two orthogonal fictitious displacements of the revolute joint $A_2$, as follows:

$$v_{10}^{Av} \vec{u}_j^T a_{10}^T \vec{u}_3 + v_{21}^{Ayv} \vec{u}_j^T a_{10}^T \vec{u}_2 + v_{21}^{Azv} \vec{u}_j^T a_{10}^T \vec{u}_3 + \omega_{21}^{Av} \vec{u}_j^T a_{20}^T \tilde{u}_3 \{\vec{r}_{32}^A + a_{32}^T \vec{r}_3^{GA}\} + v_{32}^{Av} \vec{u}_j^T a_{30}^T \vec{u}_3 = \vec{u}_j^T \vec{v}_0^{Gv} \quad (j = 1, 2). \tag{15}$$

Now, let us assume that the robot has successively some virtual motions determined by following sets of velocities:

$$v_{10a}^{Av} = 1, \ v_{10a}^{Bv} = 0, \ v_{10a}^{Cv} = 0, \ v_{21a}^{iyv} = 0, \ v_{21a}^{izv} = 0$$

$$v_{10a}^{iv} = 0, \ v_{21a}^{Ayv} = 1, \ v_{21a}^{Byv} = 0, \ v_{21a}^{Cyv} = 0, \ v_{21a}^{izv} = 0$$

$$v_{10a}^{iv} = 0, \ v_{21a}^{iyv} = 0, \ v_{21a}^{Azv} = 1, \ v_{21a}^{Bzv} = 0, \ v_{21a}^{Czv} = 0 \quad (i = A, B, C). \quad (16)$$

These virtual velocities are required into the computation of virtual power and virtual work of all forces applied to the component elements of the robot.

As for the relative accelerations $\gamma_{10}^A, \varepsilon_{21}^A, \gamma_{32}^A$ of the robot, new conditions of connectivity are obtained by the derivative of above equations (14):

$$\gamma_{10}^A \vec{u}_j^T a_{10}^T \vec{u}_3 + \gamma_{32}^A \vec{u}_j^T a_{30}^T \vec{u}_3 = \vec{u}_j^T \ddot{\vec{r}}_0^G - \dot{\phi}^2 \vec{u}_j^T a_{20} \tilde{\vec{u}}_3 \tilde{\vec{u}}_3 \{\vec{r}_{32}^A + a_{32}^T \vec{r}_3^{GA}\} - \ddot{\phi} \vec{u}_j^T a_{20} \tilde{\vec{u}}_3 \{\vec{r}_{32}^A + a_{32}^T \vec{r}_3^{GA}\} -$$
$$- 2 v_{32}^A \dot{\phi} \vec{u}_j^T a_{20} \tilde{\vec{u}}_3 a_{32}^T \vec{u}_3 \quad (j = 1, 2). \quad (17)$$

The following recursive relations give the angular accelerations $\vec{\varepsilon}_{k0}^A$ and the accelerations $\vec{\gamma}_{k0}^A$ of joints $A_k$

$$\vec{\gamma}_{10}^A = \ddot{\lambda}_{10}^A \vec{u}_3, \ \vec{\gamma}_{21}^A = \vec{0}, \ \vec{\gamma}_{32}^A = \ddot{\lambda}_{32}^A \vec{u}_3, \ \vec{\varepsilon}_{10}^A = \vec{0}, \ \vec{\varepsilon}_{21}^A = \ddot{\phi} \vec{u}_3, \ \vec{\varepsilon}_{32}^A = \vec{0}$$

$$\vec{\varepsilon}_{k0}^A = a_{k,k-1} \vec{\varepsilon}_{k-1,0}^A + \varepsilon_{k,k-1}^A \vec{u}_3 + \omega_{k,k-1}^A a_{k,k-1} \tilde{\vec{\omega}}_{k-1,0}^A a_{k,k-1}^T \vec{u}_3$$

$$\tilde{\vec{\omega}}_{k0}^A \tilde{\vec{\omega}}_{k0}^A + \tilde{\vec{\varepsilon}}_{k0}^A = a_{k,k-1}(\tilde{\vec{\omega}}_{k-1,0}^A \tilde{\vec{\omega}}_{k-1,0}^A + \tilde{\vec{\varepsilon}}_{k-1,0}^A) a_{k,k-1}^T + \omega_{k,k-1}^A \omega_{k,k-1}^A \tilde{\vec{u}}_3 \tilde{\vec{u}}_3 + \varepsilon_{k,k-1}^A \tilde{\vec{u}}_3 + 2 \omega_{k,k-1}^A a_{k,k-1} \tilde{\vec{\omega}}_{k-1,0}^A a_{k,k-1}^T \tilde{\vec{u}}_3$$

$$\vec{\gamma}_{k0}^A = a_{k,k-1} \vec{\gamma}_{k-1,0}^A + a_{k,k-1}(\tilde{\vec{\omega}}_{k-1,0}^A \tilde{\vec{\omega}}_{k-1,0}^A + \tilde{\vec{\varepsilon}}_{k-1,0}^A) \vec{r}_{k,k-1}^A + 2 v_{k,k-1}^A a_{k,k-1} \tilde{\vec{\omega}}_{k-1,0}^A a_{k,k-1}^T \vec{u}_3 + \vec{\gamma}_{k,k-1}^A \vec{u}_3, \ (k = 1, 2, 3). \quad (18)$$

## 3. DYNAMICS EQUATIONS

The dynamics analysis of parallel robots is complicated because the existence of a spatial kinematical structure, which possesses a large number of passive degrees of freedom, dominance of the inertial forces, frictional and gravitational components and by the problem linked to real-time control in the inverse dynamics. In the context of the real-time control, neglecting the frictions forces and considering the gravitational effects, the relevant objective of the complete dynamics is first to determine the input torques or forces, which must be exerted by the actuators in order to produce a given trajectory of the end-effector, but also to calculate all internal joint forces or torques.

A lot of works have focused on the dynamics of Stewart platform. Dasgupta and Mruthyunjaya [18] used the Newton-Euler approach to develop closed-form dynamic equations of Stewart platform, considering all dynamic and gravity effects as well as viscous friction at joints. Tsai [1] presented an algorithm to solve the inverse dynamics for a Stewart platform-type using Newton-Euler equations. This commonly known approach requires computation of all constraint forces and moments between the links.

Three independent mechanical systems acting along the planar directions $A_1 z_1^A$, $B_1 z_1^B$ and $C_1 z_1^C$ with the forces $\vec{f}_{10}^A = f_{10}^A \vec{u}_1$, $\vec{f}_{10}^B = f_{10}^B \vec{u}_1$, $\vec{f}_{10}^C = f_{10}^C \vec{u}_1$ can control the general motion of the moving platform. Knowing the position and kinematics state of each link as well as the external forces acting on the 3-*PRP* planar parallel robot, in the present paper we apply the principle of virtual work for the inverse dynamic problem in order to establish some definitive recursive matrix relations for the calculus of input torques of the actuators and internal forces in the joints. The parallel robot can artificially be transformed in a set of three open chains $C_i$ $(i = A, B, C)$ subject to the constraints. This is possible by cutting each joint for moving platform, and takes its effect into account by introducing the corresponding constraint conditions. The first and more complicated open tree system includes the first acting link and comprises also the moving platform.

The wrench of two vectors $\vec{F}_k^*$ and $\vec{M}_k^*$ evaluates the influence of the action of the weight $m_k \vec{g}$ and of other external and internal forces applied to the same element $T_k$ of the mechanism

$$\vec{F}_k^* = 9.81 m_k q_{k0} \vec{u}_2, \ \vec{M}_k^* = 9.81 m_k \tilde{\vec{r}}_k^C q_{k0} \vec{u}_2, \ (q = a, b, c). \quad (19)$$

Now, we compute the force of inertia $\vec{F}_k^{in}$ and the resulting moment of inertia forces $\vec{M}_k^{in}$ of an arbitrary rigid body $T_k$ of mass $m_k$ with respect to the centre of its first joint:

$$\vec{F}_k^{\,in} = -m_k\{\vec{\gamma}_{k0} + (\tilde{\omega}_{k0}^2 + \tilde{\varepsilon}_{k0})\vec{r}_k^{\,C}\}, \quad \vec{M}_k^{\,in} = -m_k\{\tilde{r}_k^{\,C}\vec{\gamma}_{k0} + \hat{J}_k\vec{\varepsilon}_{k0} + \tilde{\omega}_{k0}\hat{J}_k\vec{\omega}_{k0}\}. \qquad (20)$$

Pursuing the first leg *A*, for example, two significant recursive relations generate the vectors

$$\vec{F}_k^{\,A} = \vec{F}_{k0}^{\,A} + a_{k+1,k}^T \vec{F}_{k+1}^{\,A}, \quad \vec{M}_k^{\,A} = \vec{M}_{k0}^{\,A} + a_{k+1,k}^T \vec{M}_{k+1}^{\,A} + \tilde{r}_{k+1,k}^{\,A} a_{k+1,k}^T \vec{F}_{k+1}^{\,A} \qquad (21)$$

where one denoted

$$\vec{F}_{k0}^{\,A} = -\vec{F}_k^{\,inA} - \vec{F}_k^{\,*A}, \quad \vec{M}_{k0}^{\,A} = -\vec{M}_k^{\,inA} - \vec{M}_k^{\,*A}. \qquad (22)$$

Considering some independent virtual motions of the robot, all virtual displacements and velocities should be compatible with the virtual motions imposed by all kinematical constraints and joints at a given instant in time. The fundamental principle of the virtual work states that a mechanism is under dynamic equilibrium if and only if the virtual work developed by all external, internal and inertia forces vanish during any general virtual displacement, which is compatible with the constraints imposed on the mechanism. Assuming that frictional forces at the joints are negligible, the virtual work produced by all remaining forces of constraint at the joints is zero.

Total virtual work contributed by the inertia forces and moments of inertia forces $\vec{F}_k^{\,in}, \vec{M}_k^{\,in}$, by the wrench of known external forces $\vec{F}_k^{\,*}, \vec{M}_k^{\,*}$ and by the first active force $\vec{f}_{10}^{\,A}$ or some internal joint forces, for example, can be written in a compact form, based on the relative virtual velocities. Applying *the fundamental equations of the parallel robots dynamics* [19], the following compact matrix relations results

$$f_{10}^{\,A} = \vec{u}_3^T \vec{F}_1^{\,A} + \vec{u}_3^T\{\omega_{21a}^{Av}\vec{M}_2^{\,A} + v_{32a}^{Av}\vec{F}_3^{\,A} + \omega_{21a}^{Bv}\vec{M}_2^{\,B} + \omega_{21a}^{Cv}\vec{M}_2^{\,C}\} \qquad (23)$$

for the *input force* of first prismatic actuator,

$$f_{21y}^{\,A} = \vec{u}_2^T a_{21}^T \vec{F}_2^{\,A} + \vec{u}_3^T\{\omega_{21a}^{Av}\vec{M}_2^{\,A} + v_{32a}^{Av}\vec{F}_3^{\,A} + \omega_{21}^{Bv}\vec{M}_2^{\,B} + \omega_{21}^{Cv}\vec{M}_2^{\,C}\} \qquad (24)$$

for the first *joint force* and

$$f_{21z}^{\,A} = \vec{u}_3^T a_{21}^T \vec{F}_2^{\,A} + \vec{u}_3^T\{\omega_{21a}^{Av}\vec{M}_2^{\,A} + v_{32a}^{Av}\vec{F}_3^{\,A} + \omega_{21}^{Bv}\vec{M}_2^{\,B} + \omega_{21}^{Cv}\vec{M}_2^{\,C}\} \qquad (25)$$

for the second *joint force* acting in the joint $A_2$. The relations (21) - (25) represent the *complete inverse dynamics model* of the *3-PRP* planar parallel robot.

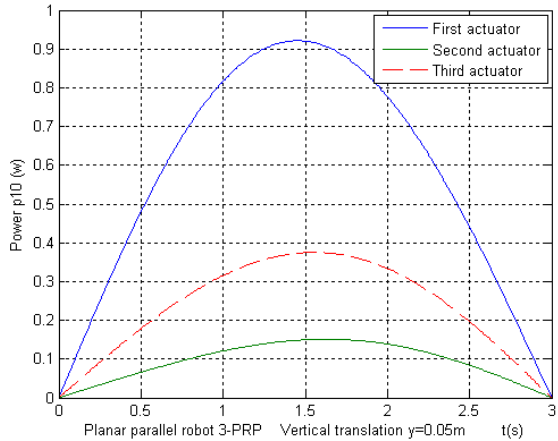
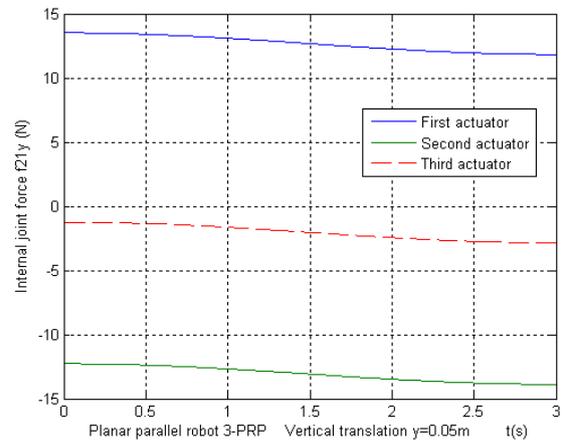

Fig. 3 Powers $p_{10}^A, p_{10}^B, p_{10}^C$ of three actuators　　　Fig. 4 Internal joint forces $f_{21y}^A, f_{21y}^B, f_{21y}^C$

As application let us consider a *3-PRP* planar robot, which has the following geometrical and architectural characteristics:

$$x_0^{G*} = 0, \ y_0^{G*} = 0.025\,m, \ \phi^* = \frac{\pi}{12}, \ l_0 = OA_0 = OB_0 = OC_0 = 0.3m, \ l = l_0\sqrt{3}$$

$$m_1 = 1\,kg, \ m_2 = 0.75\,kg, \ m_3 = 3\,kg, \ \Delta t = 3\,s.$$

Assuming that there are no external forces and moments acting on the moving platform, a dynamic simulation is based on the computation of three powers $p_{10}^i$ required by each actuator during the platform's evolution and the internal joint forces $f_{21y}^i, f_{21z}^i$ $(i = A, B, C)$. Using the MATLAB software, a computer program was developed to solve the inverse dynamics of the planar parallel robot. To illustrate the algorithm, it is assumed that for a period of three second the platform starts at rest from a central configuration and rotates or moves along rectilinear directions.

If the platform's centre $G$ moves along a *vertical trajectory* without rotation of platform, the input powers of three actuators and some joint forces are calculated by the program and plotted versus time as follows: Fig. 3 - Fig. 5. For the second example we consider the *rotation motion* of the moving platform about $z_0$ horizontal axis with variable angular acceleration while all the other positional parameters are held equal to zero (Fig. 6 - Fig. 8).

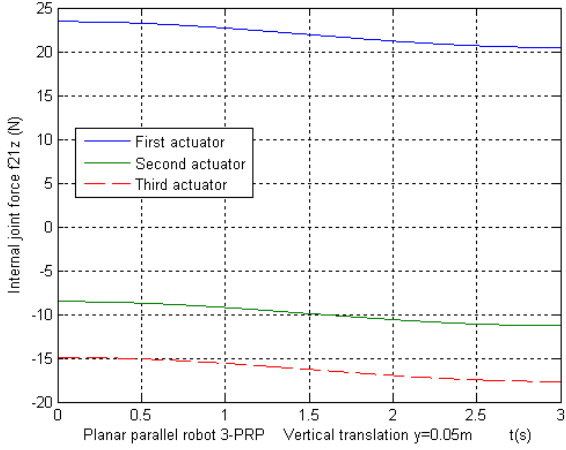

Fig. 5 Internal joint forces $f_{21z}^A, f_{21z}^B, f_{21z}^C$

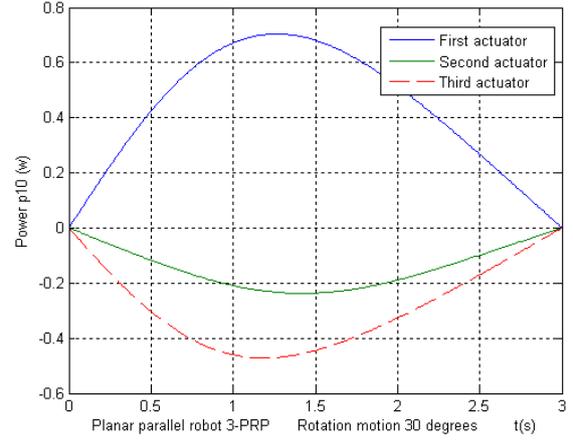

Fig. 6 Powers $p_{10}^A, p_{10}^B, p_{10}^C$ of three actuators

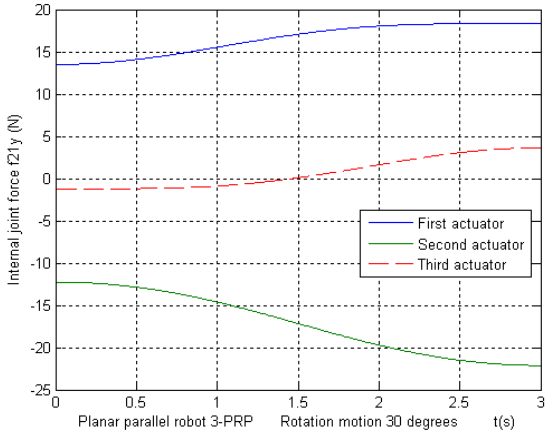

Fig. 7 Internal joint forces $f_{21y}^A, f_{21y}^B, f_{21y}^C$

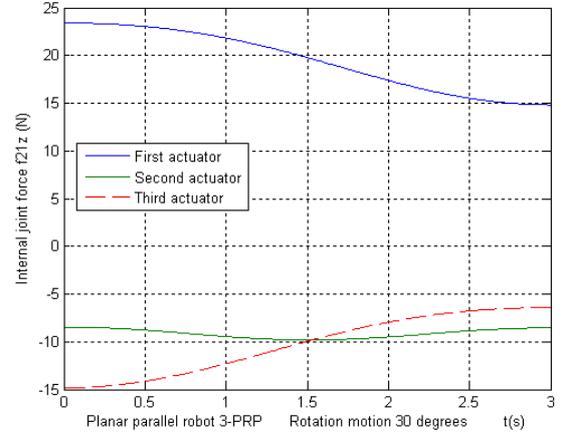

Fig. 8 Internal joint forces $f_{21z}^A, f_{21z}^B, f_{21z}^C$

The simulation through the MATLAB program certify that one of the major advantages of the current matrix recursive formulation is a reduced number of additions or multiplications and consequently a smaller processing time of numerical computation.

## 4. CONCLUSIONS

Within the inverse kinematics analysis some exact relations that give in real-time the position, velocity and acceleration of each element of the parallel robot have been established in the present paper. The dynamics model takes into consideration the mass, the tensor of inertia and the action of weight and inertia force introduced by all compounding elements of the parallel mechanism.

Based on the principle of virtual work, this approach can eliminate all forces of internal joints and establishes a direct determination of the time-history evolution for the torques required by the actuators and the internal forces or torques in joints. Choosing appropriate serial kinematical circuits connecting many moving platforms, the present method can easily be applied in forward and inverse mechanics of various types of parallel mechanisms, complex manipulators of higher degrees of freedom and particularly *hybrid structures*, when the number of components of the mechanisms is increased.